\pdfoutput=1
\documentclass[11pt]{article}

\usepackage{ACL2023}

\usepackage{times}
\usepackage{latexsym}
\usepackage{graphicx} 
\usepackage{booktabs}
\usepackage{tabularx}
\usepackage{url}
\usepackage{amsmath}
\usepackage{amssymb}
\usepackage{enumitem}
\usepackage{multirow} 
\usepackage{hyperref} 
\usepackage{xcolor}
\usepackage{pifont}
\usepackage{colortbl}
\usepackage{framed}
\definecolor{darkred}{RGB}{255,150,150}     % for values < 0.02
\definecolor{lightred}{RGB}{255,200,200}    % for values 0.02-0.03
\definecolor{darkyellow}{RGB}{255,230,150}  % for values 0.03-0.05
\definecolor{lightyellow}{RGB}{255,255,200} % for values 0.05-0.10
\definecolor{lightgreen}{RGB}{220,255,220}  % for values 0.10-0.15
\definecolor{darkgreen}{RGB}{180,255,180}   % for values > 0.15

\usepackage[T1]{fontenc}
\usepackage[utf8]{inputenc}

\usepackage{microtype}

\usepackage{inconsolata}

\title{\textsc{Decisive}: Guiding User Decisions with Optimal Preference Elicitation from Unstructured Documents}

\author{
    Akriti Jain$^{1*}$ \quad 
    Anish Mulay$^{2*\dagger}$ \quad 
    Divyansh Verma$^{3*\dagger}$ \quad 
    Aishani Pandey$^{4\dagger}$ \\
    \textbf{Pritika Ramu}$^{5\ddagger}$ \quad 
    \textbf{Aparna Garimella}$^{1}$ \\[0.5em]
    $^1$Adobe Research, India \quad $^2$IIT Madras \quad $^3$IIT Roorkee \\
    $^4$IIIT Hyderabad \quad $^5$University of Maryland, College Park \\
    \texttt{\{akritij, garimell\}@adobe.com}
}
\begin{document}
\maketitle
\begin{abstract}
\def\thefootnote{*}\footnotetext{Equal Contribution}\def\thefootnote{\arabic{footnote}}
\def\thefootnote{†}\footnotetext{Work done during internship at Adobe Research}\def\thefootnote{\arabic{footnote}}
\def\thefootnote{‡}\footnotetext{Work done while at Adobe Research}\def\thefootnote{\arabic{footnote}}

% \def\thefootnote{*}
% \footnotetext{Equal Contribution}

% \def\thefootnote{\dagger}
% \footnotetext{Work done during internship at Adobe Research}

\def\thefootnote{\arabic{footnote}}

% \author{First Author \\
%   Affiliation / Address line 1 \\
%   Affiliation / Address line 2 \\
%   Affiliation / Address line 3 \\
%   \texttt{email@domain} \\\And
%   Second Author \\
%   Affiliation / Address line 1 \\
%   Affiliation / Address line 2 \\
%   Affiliation / Address line 3 \\
%   \texttt{email@domain} \\}

% \begin{document}
% \maketitle
% \begin{abstract}

Decision-making is a cognitively intensive task that requires synthesizing relevant information from multiple unstructured sources, weighing competing factors, and incorporating subjective user preferences. Existing methods, including large language models and traditional decision-support systems, fall short: they often overwhelm users with information or fail to capture nuanced preferences accurately. We present \textbf{\textsc{Decisive}}, an interactive decision-making framework that combines document-grounded reasoning with Bayesian preference inference. Our approach grounds decisions in an objective option-scoring matrix extracted from source documents, while actively learning a user's latent preference vector through targeted elicitation. Users answer pairwise tradeoff questions adaptively selected to maximize information gain over the final decision. This process converges efficiently, minimizing user effort while ensuring recommendations remain transparent and personalized. Through extensive experiments, we demonstrate that our approach significantly outperforms both general-purpose LLMs and existing decision-making frameworks achieving up to \textbf{20\%} improvement in decision accuracy over strong baselines across domains.
% Decision-making is a cognitively intensive task that requires synthesizing relevant information from multiple unstructured evidence sources, weighing competing factors, and incorporating subjective user preferences. Existing methods, including large language models and traditional decision-support systems, fall short: they overwhelm users with information, fail to capture nuanced preferences, and provide little transparency into their recommendations. We present \textbf{\textsc{Decisive}}, an interactive decision-making framework that combines document-grounded reasoning with probabilistic preference modeling to overcome these limitations. Our method first extracts preference-based factors from source documents and organizes them into a structured framework to score each decision option. To personalize recommendations, it employs an active elicitation strategy in which users answer pairwise tradeoff questions adaptively selected to maximize expected information gain. User responses trigger Bayesian belief updates, progressively refining a utility distribution over decision options. This process converges efficiently, minimizing user effort while ensuring recommendations remain transparent and personalised. Through extensive experiments, we demonstrate the superior performance of our proposed approach over both general-purpose LLMs and existing end-to-end decision-making systems.
\end{abstract}

\section{Introduction}

% \AJ{1. Mention pain-points of users in making decisions, motivate use of AI 2. Illustrate problems in AI-assisted decision making 3. Mention our contribution}
% \AJ{highlight why LLMs -- there are works you can cite for this}
% \AJ{Also why human intervention}
% \AJ{There are three options, either user does it themselves (most accurate but most time-taking) vs GPT baseline (least accurate and least time-taking). Our system falls in between these two where we have the accuracy and time in between}

\begin{figure}[t!]
    \centering
    \includegraphics[width=0.8\linewidth]{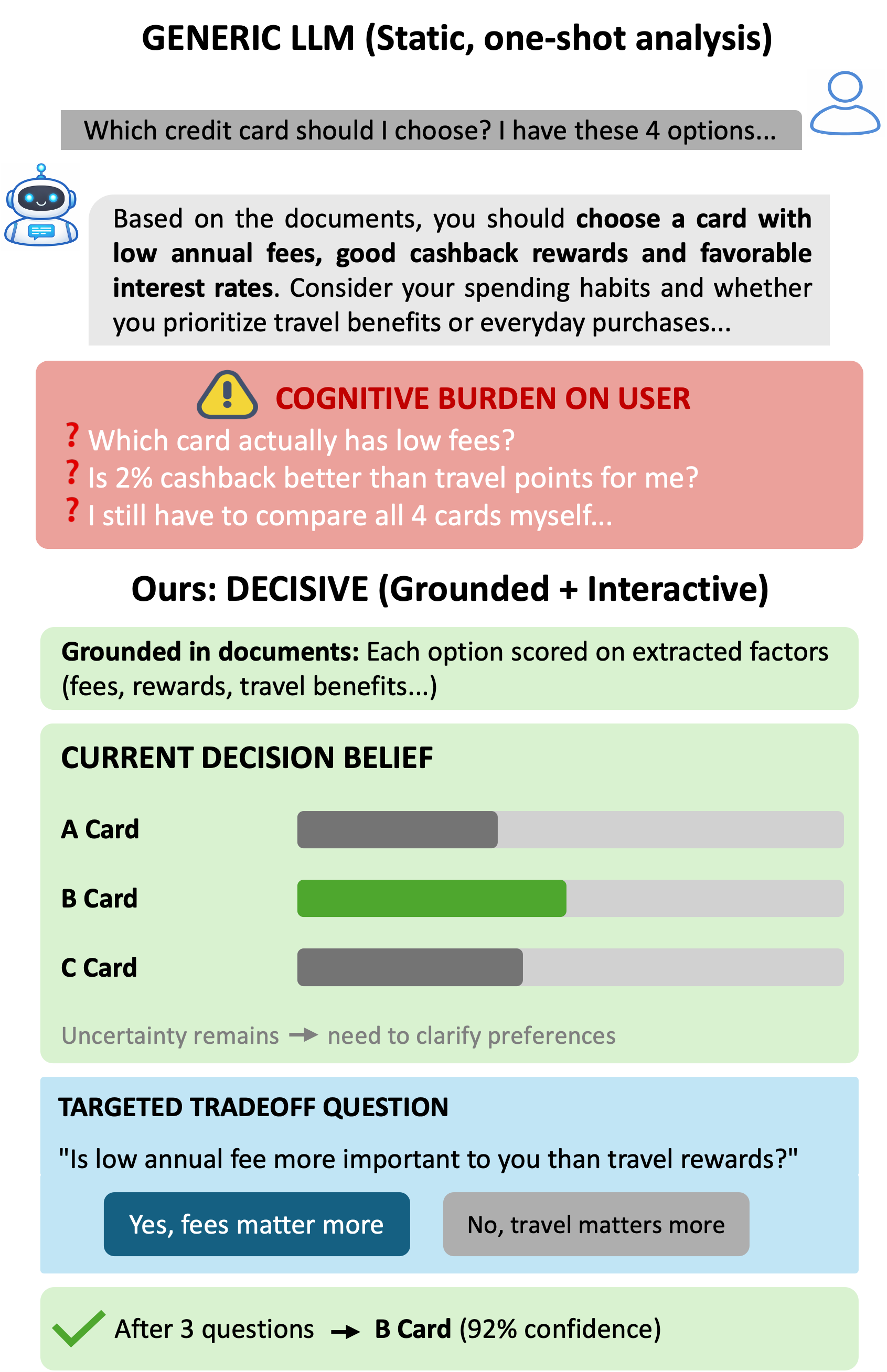}
        \caption{Comparison of decision support approaches. \textbf{Top}: A generic LLM provides abstract advice (e.g., ``choose a plan with low fees and high rewards''), forcing the user to manually verify which option fits. \textbf{Bottom}: \textsc{Decisive} grounds recommendations in document-extracted scores and actively elicits preferences through targeted tradeoff questions to find the optimal choice.}
    \label{fig:intro_comparison}
\end{figure}

Making informed decisions in high-stakes domains, such as selecting a suitable credit card, choosing an educational program, or defining a corporate strategy, requires synthesizing vast amounts of unstructured information from diverse document collections (e.g., financial disclosures, user reviews, technical reports). These decisions are inherently multi-objective, demanding that users weigh competing factors like cost versus quality or short-term gain versus long-term stability. While retrieving relevant information is a necessary first step, a sound decision-making process requires more than just locating passages. It requires the decision-relevant factors to be systematically extracted and consistently evaluated, enabling users to weigh trade-offs that align with their specific, often latent, goals.

% A sound decision-making process requires more than locating relevant passages; it demands that decision-relevant criteria be systematically extracted from these texts and that each option be consistently evaluated against them.
% While retrieving relevant information is a necessary first step, the core challenge lies in evaluating trade-offs that align with the user's specific, often latent, goals.
% In the past few years, we see the growing integration of AI in decision-making (e.g., clinical diagnosis, finance), with the goal of complementing human expertise, not just automating tasks. Despite this promise, achieving true human-AI collaboration is difficult. Mention that many systems are "static information providers," which can lead to problems like cognitive overload, ineffective reliance, and a failure to align with the human user's goals and values .
% Individuals and organizations frequently make critical choices, such as selecting a suitable credit card, choosing an educational program, or setting corporate strategy based on extensive and diverse document collections (e.g., financial disclosures, user reviews, technical reports). A sound decision-making process requires more than locating relevant passages; it demands that decision-relevant criteria be systematically extracted from these texts and that each option be consistently evaluated against them. These choices are inherently multi-objective, involving complex trade-offs between factors like cost and quality or short-term gains and long-term resilience.
To manage such complexity, AI-powered decision support is now widespread. In finance, models optimize investment portfolios~\cite{li2024investorbenchbenchmarkfinancialdecisionmaking}; in healthcare, they assist with clinical diagnosis~\cite{nazi2024largelanguagemodelshealthcare}; and in strategic planning, they support complex reasoning~\cite{changeux2024strategic}. 
% Traditionally, AI has acted as an assistive tool for well-defined analytical tasks like classification or summarization. 
Large Language Models (LLMs) have transformed this landscape, offering powerful capabilities to process the vast quantities of unstructured text central to these decisions. Yet despite this promise, current LLM-based solutions face significant limitations. Off-the-shelf LLMs typically act as ``static information providers'', delivering generic, one-shot analyses that fail to incorporate the user's unique context or latent preferences (Fig.\ref{fig:intro_comparison}). Consequently, the cognitive burden of synthesis and alignment is forced back onto the human, who must interpret the AI's output and manually map it to their personal objectives. This not only negates potential efficiency gains but also fosters mistrust in systems that remain opaque and misaligned with user goals. The failure to actively model user preferences is a critical gap; indeed, recent work identifies the user's mental model as a key determinant for improving both the quality of and reliance on LLM-assisted decisions~\cite{eigner2024determinantsllmassisteddecisionmaking}.
Conversely, recent frameworks like DeLLMa~\cite{liu2024dellmadecisionmakinguncertainty} and DecisionFlow~\cite{chen2025decisionflowadvancinglargelanguage} assume idealized scenarios where all relevant information is explicitly stated and a single objective governs the choice. This overlooks two critical realities of complex decision-making: (1) evidence is often scattered across multiple unstructured documents, and (2) the optimal choice depends heavily on user-specific constraints (e.g., budget limits or career goals) that are frequently latent and must be actively elicited. We argue that effective decision support must explicitly decouple two distinct problems. First, \textbf{factual evaluation}, where options are objectively scored against decision-relevant factors extracted from documents, a process that is constant across users. Second, \textbf{preference modeling}, where the system actively learns the user's subjective priorities, a process unique to each individual. 
By separating objective evidence from subjective goals, we can build systems that are both robust and highly personalized. 

To this end, we introduce \textsc{Decisive}, an interactive framework that combines document-grounded reasoning with Bayesian preference inference. \textsc{Decisive} first extracts an objective Option-Scoring Matrix from unstructured documents. It then employs an active elicitation strategy, asking targeted tradeoff questions to refine a probabilistic model of the user's preferences. This allows the system to converge on a high-utility recommendation. To summarize, our primary contributions are:
\begin{enumerate}[noitemsep,topsep=0pt,leftmargin=*]
    \item We introduce \textsc{Decisive}, a framework that decouples decision-making into objective document grounding (via an Option-Scoring Matrix) and subjective preference modeling. This separation allows for robust handling of multi-objective trade-offs.
    \item We propose a decision-aware active elicitation strategy that selects questions to maximize information gain over the final decision, ensuring efficient convergence with minimal user effort.
    \item We curate a challenging, realistic benchmark spanning Finance, Education, and Hiring domains, requiring synthesis from unstructured documents. Extensive experiments show that \textsc{Decisive} significantly outperforms strong LLM baselines and structured frameworks in both accuracy and efficiency.
\end{enumerate}

\section{Related Works}

\subsection{Decision-making using LLMs}

Large Language Models (LLMs) have demonstrated significant potential as decision support tools across diverse fields like business \cite{changeux2024strategic,simchilevi2025largelanguagemodelssupply}, finance \cite
{li2024investorbenchbenchmarkfinancialdecisionmaking,yu2024finconsynthesizedllmmultiagent}, and medicine \cite
{nazi2024largelanguagemodelshealthcare, LI20251,maity2025llmhealthcare, 
kim2024mdagentsadaptivecollaborationllms,GUMILAR20244019, 10.1145/3715336.3735779}. However, research indicates that direct prompting often yields poor results as problem complexity increases, as models may fixate on specific information without adequately balancing evidence or aligning with user goals. To address these limitations, several structured frameworks have been proposed to ground LLM reasoning in formal theory.

One prominent direction involves integrating decision theory and utility maximization. The DeLLMa framework~\cite{liu2024dellmadecisionmakinguncertainty} guides LLMs through a multi-step process involving state enumeration, probabilistic forecasting, and utility elicitation to identify decisions that maximize expected utility. Building on this, DecisionFlow~\cite{chen2025decisionflowadvancinglargelanguage} transforms natural language scenarios into structured representations of actions, attributes, and constraints, inferring a latent utility function in a transparent manner. 
While these frameworks represent a significant step forward, they predominantly operate in idealized scenarios with explicit information and single objectives (e.g., profit maximization). As discussed earlier, this simplifies the dual challenge of real-world decision-making and highlights the need for systems that can both navigate complex information landscapes and actively elicit the nuanced, latent preferences necessary to resolve these trade-offs.

The interactive and collaborative nature of decision-making is also a critical area of exploration. Systems like ChoiceMates~\cite{park2025choicematessupportingunfamiliaronline} use multi-agent conversational interactions to help users discover diverse perspectives and construct personalized preference spaces. They emphasize keeping the user in the loop to prevent the loss of agency that often accompanies full automation. However, purely conversational approaches often struggle to provide precise, opinionated recommendations, yielding vague answers that fail to satisfy specific user needs~\cite{park2025choicematessupportingunfamiliaronline}. Without explicit preference modeling, these systems lack a mechanism to ensure recommendations reflect the user's priorities rather than the model's implicit tendencies~\cite{jia2024decisionmakingbehaviorevaluationframework}. \textsc{Decisive} addresses these challenges by grounding decisions in real-world unstructured documents (for precision) while keeping the user in the loop to elicit latent preferences (for alignment).

\subsection{Preference Elicitation and Clarification Questions}
% \AJ{Should we say - our work builds on this probabilistic foundation + basis for trade-offs?}
The ability of LLMs to resolve ambiguity through interaction is well-studied in literature. Early work focused on reactive clarification, where models were trained or prompted to identify underspecified queries and ask follow-up questions to resolve implicit assumptions~\cite{zhang-choi-2025-clarify, chang_teaching_2025}. While effective for simple disambiguation, these approaches often struggle in complex decision-making scenarios where the missing information (the user's latent preference structure) is abstract and high-dimensional.

Recent research has shifted towards proactive information gathering. Several benchmarks, such as InfoQuest~\cite{deoliveira2025infoquestevaluatingmultiturndialogue} and latent preference discovery tasks~\cite{tsaknakis2025llmsrecognizelatentpreferences}, have highlighted that off-the-shelf LLMs are inefficient at this process. They often fail to select the most informative questions, leading to long, unfocused interactions that result in preference dilution rather than convergence. To address this, newer methods employ specialized training paradigms ~\cite{andukuri2024stargateteachinglanguagemodels, dou2025togateclarifyingquestionssummarizing} that use self-improvement and trajectory optimization to reward effective questioning strategies, while other works leverage diffusion-inspired denoising via sequential funnel questions~\cite{montazeralghaem2025askingclarifyingquestionspreference}.
% \AJ{However, these works are focused on answering given open-ended questions, and not for complex decision-making scenarios ...}
However, these approaches optimize for reconstructing a general user profile rather than resolving the specific decision at hand. 
Consequently, they may spend interaction effort on preferences that do not differentiate the available options. 
% rather than resolving the specific decision at hand. Consequently, they may elicit preferences that, while accurate, have no bearing on which option is ultimately best.

 % However, these methods require extensive task-specific training data and lack explicit probabilistic grounding, making it difficult to provide theoretical guarantees on convergence or optimality.

A complementary direction, most relevant to our work, augments LLMs with explicit probabilistic reasoning. Foundational work in Bayesian preference elicitation~\cite{pmlr-v9-guo10b} established multi-attribute preference learning with pairwise comparison queries, and recent methods have built on this by combining LLMs with information-theoretic objectives: Active Preference Inference~\cite{piriyakulkij2024activepreferenceinferenceusing} selects questions that maximize entropy reduction, the OPEN framework~\cite{handa2024bayesianpreferenceelicitationlanguage} uses Bayesian Optimal Experimental Design to track user persona distributions, and~\citet{10.1145/3640457.3688142} combine Bayesian optimization with LLM-based acquisition functions for natural language elicitation. However, these methods are often limited to binary Yes/No queries and frequently default to direct option comparisons (e.g., ``Do you prefer Option A or B?''), which assumes the user already understands the trade-offs the system is meant to surface. The Multi-Attribute Decision Making literature~\cite{10.1145/1064009.1064038, 10.1609/aimag.v29i4.2200} reinforces this concern, showing that users construct preferences \textit{through} interaction rather than arriving with fixed ones, and that interactive tradeoff support can improve decision accuracy by up to 57\%. \textsc{Decisive} builds on these insights but introduces a key distinction: rather than reducing uncertainty over preferences alone, it optimizes for reducing uncertainty in the final \textit{decision}, focusing elicitation on the specific trade-offs that distinguish the top candidates.

\section{Dataset Curation}

Existing benchmarks such as MTA~\cite{hu2024languagemodelsalignabledecisionmakers} and DeLLMa~\cite{liu2024dellmadecisionmakinguncertainty} frame tasks around a single objective (e.g., maximizing profit) and provide contexts where all necessary information is explicitly stated. This sidesteps two core difficulties of real-world decisions: (1) locating relevant details scattered across multiple documents, and (2) weighing competing attributes (e.g., a university's research rankings or program structure) against personal constraints (e.g., tuition limits). To address this gap, we curate a dataset spanning three domains: \textbf{Finance}, \textbf{Education}, and \textbf{Hiring}. These correspond to multi-objective decisions commonly faced by knowledge workers, students, and HR professionals, respectively. As identified by prior surveys, these domains serve as representative settings for AI-assisted decision support, with applications in financial advising, educational planning, and career-related decision making~\cite{10.1145/3706598.3713423, ALBASHRAWI2025100751, yin_responsible_2025}. Each domain requires synthesizing information across multiple documents and reasoning about trade-offs that depend on user-specific goals and constraints.
\subsection{Data Sourcing}
Curating coherent document collections at scale presents significant challenges. Scraping documents from the web (e.g., tens of university brochures or loan policies) is impractical due to inconsistent formatting, access restrictions, and highly variable information granularity across sources. Instead, we adopt a hybrid sourcing strategy.
For the \textbf{Hiring} domain, we use resumes from a public Kaggle dataset~\cite{bhawal2021resume}, also used by other works \cite{veldanda2023emilygregemployablelakisha, wang-etal-2024-jobfair}. In this dataset, resumes are organized by category of occupation (e.g., engineer, consultant). For each scenario, we randomly select 10 resumes from the relevant occupational category and generate the decision question taking into account that particular role. For \textbf{Finance} and \textbf{Education}, we develop a synthetic generation pipeline grounded in manually scraped seed documents and decision scenarios. These seeds, drawn from authentic sources (e.g., real loan brochures, university prospectuses), serve as few-shot examples that anchor the generation in realistic structures and terminology.

\subsection{Synthetic Data Generation Pipeline}

\begin{figure}[t] 
    \centering
    \small
    \includegraphics[width=0.35\textwidth]{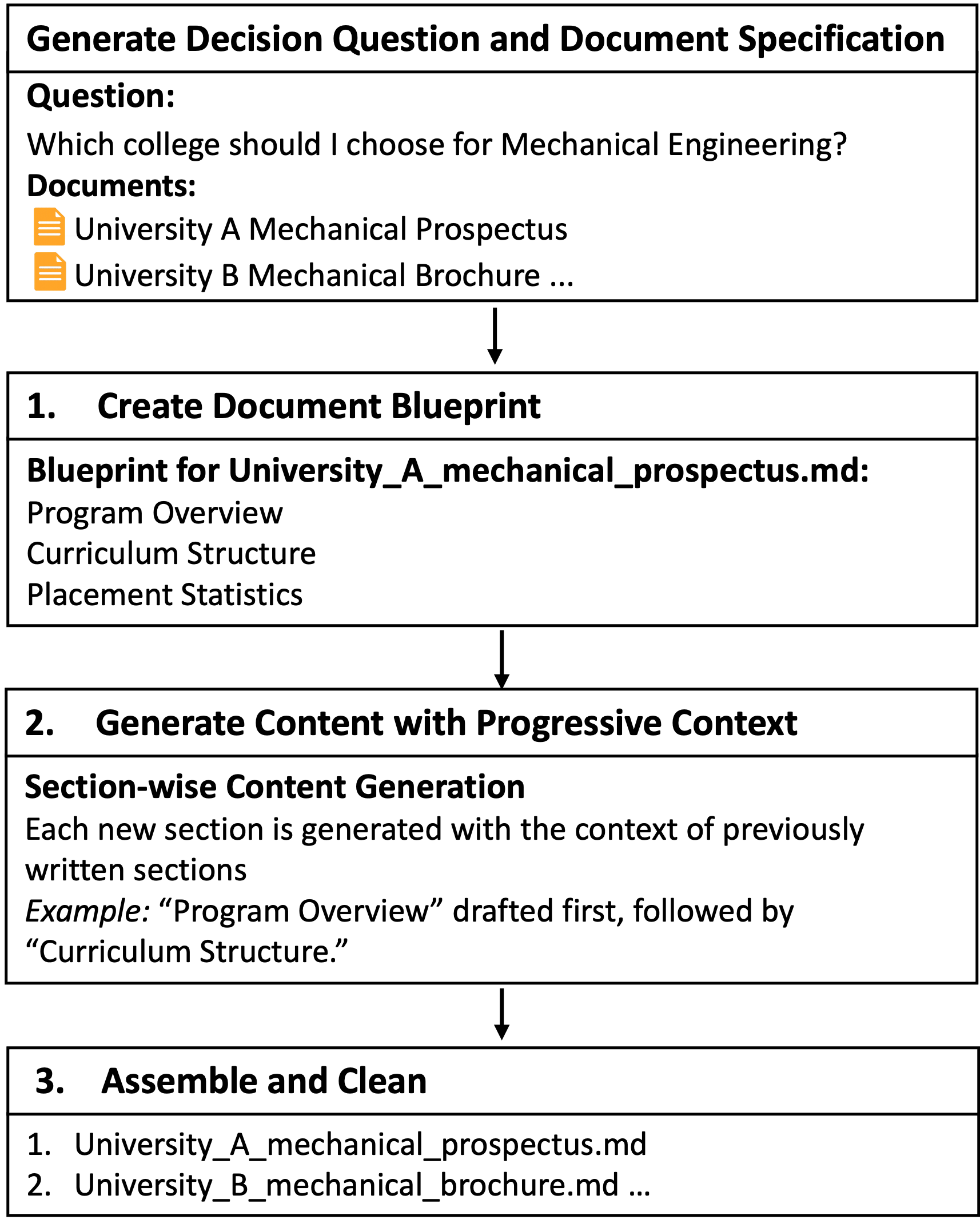} 
    \caption{Overview of the data generation process.}
    \label{fig:sample}
    \vspace{-0.1in}
\end{figure}

\paragraph{Stage 1: Question and Schema Generation.}
% \AJ{add about the seed docs part. Also we should mention about seed decisions and manual verification we conducted for that, gen 5k then filter}
We first manually curate a set of seed decision questions representative of each domain. Using these as few-shot examples, we prompt GPT-4o to generate diverse decision scenarios (e.g., ``Which graduate program best aligns with my career goals?''). A total of 500 questions are generated per domain (1,500 across three domains). For the Finance and Education domains, the model also produces a structured document schema along with the decision question, specifying the necessary documents for making an informed decision, such as program borchures etc. Each scenario includes $M$=10 candidate options, deliberately increasing the decision complexity beyond pre-existing benchmarks (which typically use 2-4 options) and ensuring that simple heuristics are insufficient. Since our preference elicitation objective operates over factors rather than options, the framework scales naturally with $M$. We further verify that performance generalizes to smaller candidate sets through an ablation at $M$=5 (Appendix~\ref{sec:appendix_m5}).
% This schema is divided into two categories:
% \begin{enumerate}[label=(\alph*),wide,labelindent=0pt, noitemsep]
%     \item \textbf{Personal Documents}: User-specific documents that provide context about the decision-maker, such as a student's academic transcript or a company's financial statement. To ensure consistency across these documents, we first generate a synthetic user/entity profile to ground the generation process.
%     \item \textbf{Option Documents}: Documents that detail the available choices, such as university course brochures, loan product descriptions, or candidate resumes.
% \end{enumerate}

\paragraph{Stage 2: Context-Aware Document Synthesis.}
% \AJ{optimising for consistency; conditioned generation, no need to verify we specificaly include that}

% The second stage synthesizes the documents defined in the schema. To ensure intra-document consistency, the pipeline performs a hierarchical, context-aware generation process. First, the system generates a JSON blueprint outlining the document's structure (sections, subsections). Content is then generated sequentially, with each section explicitly conditioned on all previously generated sections to maintain factual and stylistic coherence. Finally, a concluding LLM call compiles all sections and refines the complete document to ensure stylistic uniformity and remove generation artifacts.

The second stage synthesizes the documents defined in the schema. To ensure intra-document consistency, the pipeline performs a hierarchical, context-aware generation process: first generating a structural blueprint, then sequentially producing content where each section is conditioned on prior sections, and finally refining the compiled document for stylistic uniformity (see Appendix~\ref{sec:appendix_generation} for details). The output is a dataset where each instance consists of a decision question and a set of option documents.

% The second stage synthesizes the documents defined in the schema. To ensure intra-document consistency, the pipeline performs a hierarchical, context-aware generation process for each document: (a) Blueprint Generation: The system first generates a hierarchical JSON blueprint that outlines the document's structure, including its sections and subsections. (b) Sequential Content Generation: The content is generated section by section, where the generation of each new section is conditioned on the content of all previously generated sections. This sequential conditioning maintains factual and stylistic coherence across the document. (c) Compilation and Refinement: Finally, all generated sections are compiled, and a concluding LLM call refines the complete document to ensure stylistic uniformity and remove any generation artifacts. The output of this two-stage pipeline is a dataset where each instance consists of a decision question and a set of option documents.

% We also employ 3 domain experts from each field from Upwork to validate the realistic aspect of the decision questions along with their corresponding document sets whether they are faced with such questions in their professional lives. And we also get annotations for the parameters that they would consider in making these decisions that further help us design the prompt for the subsequent steps in our method or to come up with decision-relevant factors automatically using an LLM. After filtering, our dfinal dataset consists of 499 ed, 491, fin and 500 resume making a total of 1400.
\paragraph{Validation.}
To ensure realism, we recruit three domain experts per field from a freelancing platform,\footnote{\url{https://www.upwork.com}} to validate whether the decision questions reflect scenarios encountered in professional practice and whether the generated documents contain the information necessary to make an informed decision.  They were compensated at a rate of \$15/hour. Annotators also provide the key factors they would consider in such decisions, which informs our prompt design for automatic parameter extraction discussed in detail in Section \ref{sec:parameter_scoring}. After filtering, our final dataset comprises of 499 samples in Education, 491 in Finance, and 500 in Hiring, totaling to 1,490 decision scenarios. 
% Some examples are provided in Appendix~\ref{sec:appendix_data}.

% \paragraph{Problem Formulation.}
% The core challenge is that the user's personal preference weights, \(W = \{w_1, \dots, w_K\}\), are unknown. The goal of the system is to efficiently learn a posterior distribution over these weights through a minimal set of interactions. This learned distribution is then used to identify the option that maximizes the user's expected utility, defined as \(\sum_{j=1}^{K} w_j S_{ij}\). The methodology for achieving this is detailed in the following section.

\section{Problem Formulation}

Given a decision query $q$ and a set of candidate options $\mathcal{O} = \{o_1, \ldots, o_M\}$, where each option is described by a collection of unstructured documents $\mathcal{D}$, our goal is to identify the option $o^*$ that best aligns with the user's latent preferences. We assume the utility of an option depends on $K$ decision factors derived from the documents. We decompose the decision into two components:
% \begin{enumerate}[leftmargin=1.2em, itemsep=0pt, topsep=0pt]
%     \item \textbf{Option-Scoring Matrix ($\mathbf{S}$):} An $M \times K$ matrix where $S_{ij}$ quantifies the performance of option $i$ on factor $j$. Derived directly from source documents, this matrix transforms unstructured text into a structured, comparable representation of evidence for each option.
%     \item \textbf{User Preference Vector ($\mathbf{w}^*$):} A $K$-dimensional vector where $w^*_j$ reflects the user's subjective importance for factor $j$. This vector captures the user's unique trade-offs and priorities, which are unknown a priori and must be elicited through interaction.
% \end{enumerate}
{\bf (1)} \textbf{Option-Scoring Matrix ($\mathbf{S}$):} An $M \times K$ matrix where $S_{ij}$ quantifies the performance of option $i$ on factor $j$. Derived directly from source documents, this matrix transforms unstructured text into a structured, comparable representation of evidence for each option.
{\bf (2)} \textbf{User Preference Vector ($\mathbf{w}^*$):} A $K$-dimensional vector where $w^*_j$ reflects the user's subjective importance for factor $j$. This vector captures the user's unique trade-offs and priorities, which are unknown a priori and must be elicited through interaction.

% \item \textbf{User Preference Vector $\mathbf{w}^*$} (Subjective): A $K$-dimensional vector where $w^*_j$ represents the importance the user places on factor $j$. These weights are personal, subjective, and unknown; they must be discovered through interaction.

\noindent The recommended option is the one that maximizes the weighted utility:
\[
o^* = \arg\max_{i} \, (\mathbf{S} \mathbf{w}^*)_i
\]
% While $\mathbf{S}$ can be pre-computed, $\mathbf{w}^*$ is latent. This formulation separates factual extraction from preference modeling, allowing robust decision-making even with complex, multi-objective trade-offs.
$\mathbf{S}$ is document-derived and user-agnostic: it can be pre-computed once and remains constant across users, whereas $\mathbf{w}^*$ is latent and must be elicited. This formulation cleanly separates factual evaluation from preference modeling, allowing robust decision-making even with complex, multi-objective trade-offs. 
% Note that the system need not uniquely identify $\mathbf{w}^*$: when multiple preference vectors yield the same top-ranked option, the \textit{decision} remains identifiable even if the exact preference vector is not, and the system can terminate early.
% This decomposition distinguishes our work from prior approaches~\cite{liu2024dellmadecisionmakinguncertainty, chen2025decisionflowadvancinglargelanguage}, which often fail to robustly integrate multi-faceted user preferences with evidence from unstructured documents. By explicitly separating them, our framework can accurately handle decisions involving complex trade-offs across multiple competing factors.

\begin{figure*}[t] 
    \centering
    \includegraphics[width=0.9\textwidth]{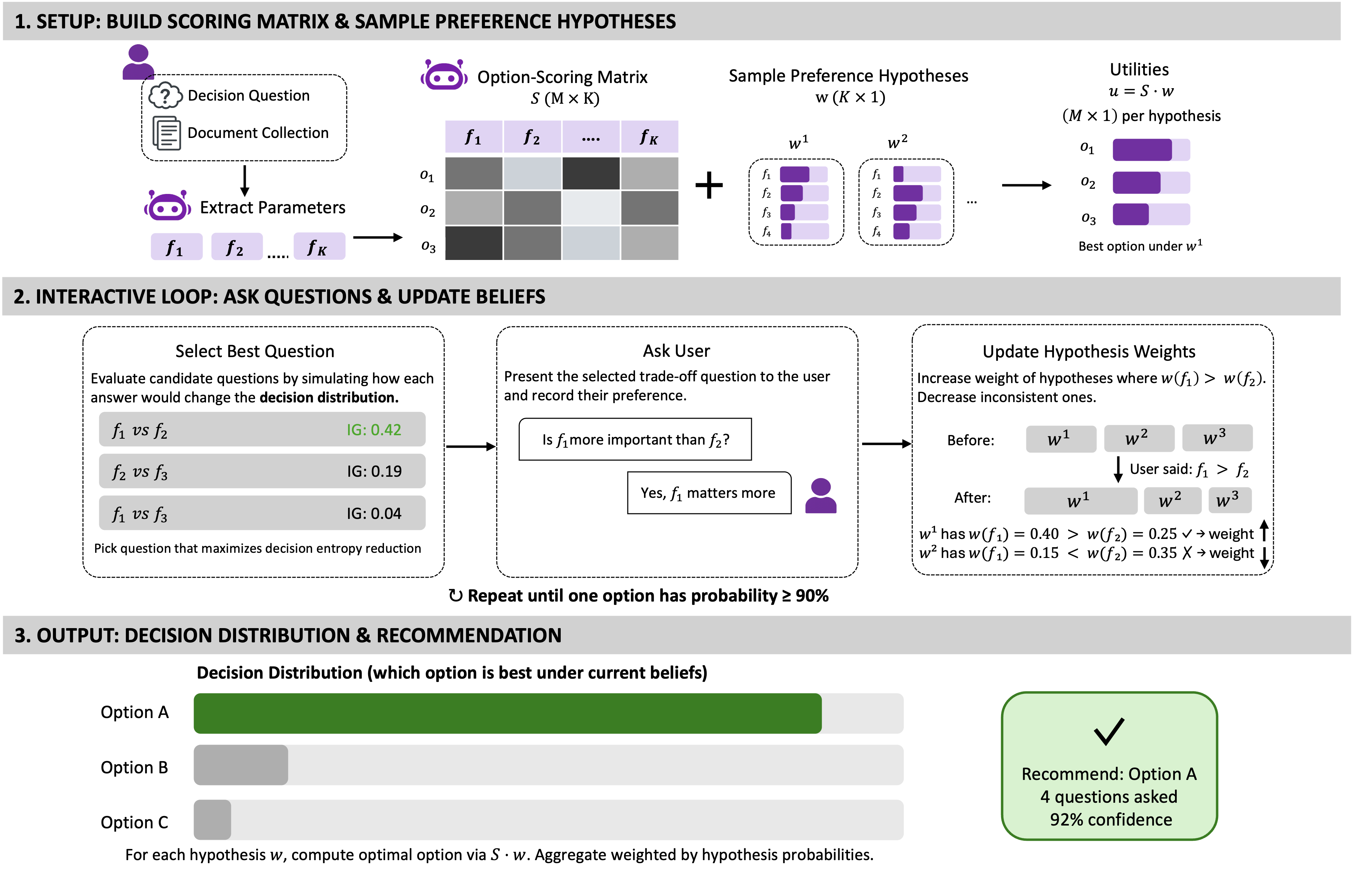} 
    \caption{Overview of the \textsc{Decisive} pipeline. The system first extracts $K$ preference factors from documents and constructs the Option-Scoring Matrix $\mathbf{S}$, which quantifies how each option performs on each factor. A Dirichlet prior models uncertainty over the user's preference weights, with particle samples representing plausible preference vectors. Decision-aware questions are adaptively selected to maximize information gain over the \textit{decision distribution}. User responses trigger Bayesian updates until decision confidence exceeds a threshold.}
    \label{fig:method}
    \vspace{-0.1in}
\end{figure*}

\section{Methodology}

Our methodology centers on a Bayesian model of user preferences that is iteratively refined using an active learning policy. This approach allows the system to intelligently explore the preference space and converge on a high-utility recommendation with minimal user effort.

\subsection{Parameter Extraction and Option-Scoring Matrix Construction}
\label{sec:parameter_scoring}
Given a decision query $q$ and documents $\mathcal{D}$, an LLM (GPT-5) first extracts $K$ decision-relevant factors $\mathcal{F}$ such as tuition cost or risk tolerance. $K$ is not fixed globally but is extracted per scenario, tailored to the specific decision question at hand. In practice, we observe an overall average of $K$=11.2 factors per scenario, with domain-level averages of 12.0 for Education (range 8-17), 13.4 for Finance (range 8-21), and 8.3 for Hiring (range 5-14).

For each option-factor pair $(o_i, f_j)$, the system generates a qualitative assessment on an 8-point ordinal scale (``Very Low'', ``Low'', ``Low to Medium'', ``Medium'', ``Medium to High'', ``High'', ``High to Very High'', ``Very High'') grounded in the source documents, which is then mapped to a numerical value in $[0,1]$ to form the matrix $\mathbf{S}$. This graded, comparative scoring is distinct from standard binary relevance judgment~\cite{upadhyay2024umbrelaumbrelaopensourcereproduction}, as our task requires nuanced relative comparison across options. To ensure robustness and mitigate individual model bias, we generate these scores using three distinct LLMs (GPT-5, Gemini-3-Pro, Claude-4.5-Sonnet) and combine them via median aggregation on the ordinal scale.

\subsection{Modeling User Preferences}

% With $\mathbf{S}$ fixed, the system must discover the user's latent preference vector $\mathbf{w}^*$. A single LLM estimate of user preferences can be unreliable. To address this, we model the uncertainty over possible preference weights by placing a symmetric Dirichlet distribution over the weight space, with $\alpha_j = 1$ to avoid initial bias. We approximate this distribution by maintaining a set of $P$ hypothesis vectors $\{\mathbf{w}^{(1)}, \ldots, \mathbf{w}^{(P)}\}$ sampled from the prior. Each vector represents a plausible user ``persona,'' initially assigned an equal likelihood weight $\pi^{(p)} = 1/P$. This represents our initial belief state about the user's preferences.

With $\mathbf{S}$ fixed, the system must discover the user's latent preference vector $\mathbf{w}^*$. A single LLM estimate of user preferences can be unreliable. To address this, we model uncertainty over preferences using a uniform Dirichlet prior ($\alpha_j = 1$), the standard uninformative prior in Bayesian inference, which assumes no initial bias toward any particular preference structure. We then sample $P$ candidate preference vectors $\{\mathbf{w}^{(1)}, \ldots, \mathbf{w}^{(P)}\}$ from this prior, each representing a plausible user ``persona'' with an equal initial likelihood weight $\pi^{(p)} = 1/P$. This particle-based approximation~\cite{warwick37961, Arulampalam2002ATO} allows us to tractably maintain and update a rich distribution over possible user preferences. This set of weighted personas represents our initial belief state about the user's preferences.

\subsection{Interactive Preference Refinement}

To refine our belief state, the system interacts with the user through simple pairwise tradeoff questions. When a user indicates that factor $a$ is preferred over factor $b$ ($f_a \succ f_b$), we perform a Bayesian update on the likelihoods of our sampled personas. The update reinforces personas consistent with the user's feedback and down-weights those that are not. Importantly, inconsistent personas are down-weighted but retained, ensuring that an occasional noisy or contradictory user response does not destroy information but simply redistributes belief across the preference space. Formally, this is achieved using a sigmoid update rule:
% \small
\[
\pi^{(p)} \leftarrow \pi^{(p)} \cdot \sigma\Big(\kappa \cdot (w^{(p)}_a - w^{(p)}_b)\Big)
\]
% \normal
where $\sigma(\cdot)$ is the sigmoid function and $\kappa$ controls the sharpness of the update. After each interaction, the likelihoods are re-normalized. This iterative process gradually concentrates the probability mass on the subset of personas most consistent with the user's cumulative responses.

% The system refines its belief state by asking pairwise tradeoff questions ($f_a$ vs. $f_b$). When a user provides feedback (e.g., $f_a \succ f_b$), we update the weight of each hypothesis based on its consistency with the response: 
% \[
% \pi^{(p)} \leftarrow \pi^{(p)} \cdot \sigma\Big(\kappa \cdot (w_a^{(p)} - w_b^{(p)})\Big)
% \]
% where $\sigma(\cdot)$ is the sigmoid function and $\kappa$ is a parameter that controls the sharpness of the update. This soft update mechanism gradually concentrates probability mass on personas that align with the user's expressed priorities.

\subsection{Decision-Aware Question Selection}

Not all questions are equally informative. To maximize efficiency, we select questions that reduce uncertainty in the \textit{final decision}, rather than just in the preference weights. We define the system's current belief as a probability distribution $\boldsymbol{\chi} \in \mathbb{R}^M$ over the available options, where $\chi_i$ represents the probability that option $i$ is optimal, marginalized over all sampled preference personas:
\[
\chi_i = \sum_{p=1}^{P} \pi^{(p)} \cdot \mathbb{I}\left[\arg\max_{i'} (\mathbf{S} \mathbf{w}^{(p)})_{i'} = i\right]
\]
We quantify the uncertainty of this distribution using its entropy $H(\boldsymbol{\chi})$. For each candidate question comparing factors $a$ and $b$, we simulate both possible user responses ($f_a \succ f_b$ and $f_b \succ f_a$) and compute the expected posterior entropy of the decision distribution. The system selects the question that maximizes the Expected Information Gain (EIG), defined as the reduction in decision entropy:
\[
\text{EIG}(a, b) = H(\boldsymbol{\chi}) - \mathbb{E}_{resp} [H(\boldsymbol{\chi}^{(resp)})]
\]
This formulation ensures the system focuses only on preference trade-offs that actually differentiate the top contenders, effectively ignoring factors that do not impact the final recommendation.

\subsection{Stopping and Recommendation}

Instead of asking a fixed number of questions, we employ a dynamic stopping criterion based on decision stability. The dialogue terminates when the confidence in the most likely option exceeds a threshold ($\max_i \chi_i \geq \tau$, with $\tau=0.85$). Upon stopping, it recommends the option with the highest expected utility under the final posterior belief:
\[
o^* = \arg\max_i \sum_{p=1}^{P} \pi^{(p)}_{\text{final}} \cdot (\mathbf{S} \mathbf{w}^{(p)})_i
\]
This ensures the final recommendation is grounded in both the objective evidence from the documents and the user's interactively refined preferences.

\begin{table*}[t]
\centering
\small
\scalebox{0.9}{
\begin{tabular}{l c c c c c c}
\toprule
\textbf{Method} & 
\textbf{Dialogue?} &
\textbf{Top-1 Acc.} & 
\textbf{Top-2 Acc.} & 
\textbf{NDCG@3} & 
\textbf{MRR} &
\textbf{Avg. Qs} \\
\midrule
\rowcolor{gray!15} \multicolumn{7}{c}{\textbf{Education Domain}} \\
\midrule
\multicolumn{7}{l}{\textsc{Decision Frameworks}} \\
DeLLMa & \ding{55} & 28.1 & 44.3 & 0.733 & 0.480 & -- \\
DecisionFlow & \ding{55} & 20.2 & 34.3 & 0.706 & 0.403 & -- \\
Active Pref. Inference & \ding{51} & 13.0 & 25.9 & 0.679 & 0.336 & 9.7 \\
\midrule
\multicolumn{7}{l}{\textsc{Prompting Baselines}} \\
LLM-Direct (B1) & \ding{55} & 64.8 & 82.4 & 0.799 & 0.782 & -- \\
LLM-CoT (B2) & \ding{55} & \cellcolor{lightred}65.5 & \cellcolor{lightred}83.5 & \cellcolor{lightred}0.804 & \cellcolor{lightred}0.786 & -- \\
LLM-Structured Dialogue (B3) & \ding{51} & 9.4 & 22.2 & 0.225 & 0.302 & 4.7 \\
LLM-Free Dialogue (B4) & \ding{51} & 19.4 & 34.7 & 0.345 & 0.398 & 2.8 \\
\midrule
% \textbf{\textsc{Decisive} (Ours)} & \ding{51} & \cellcolor{lightgreen}72.1 & \cellcolor{lightgreen}85.8 & \cellcolor{lightgreen}0.841 & \cellcolor{lightgreen}0.827 & 4.1 \\
\textbf{\textsc{Decisive} (Ours)} & \ding{51} & \cellcolor{lightgreen}78.8 & \cellcolor{lightgreen}91.6 & \cellcolor{lightgreen}0.893 & \cellcolor{lightgreen}0.879 & 5.1 \\
\midrule
\rowcolor{gray!15} \multicolumn{7}{c}{\textbf{Finance Domain}} \\ 
\midrule
\multicolumn{7}{l}{\textsc{Decision Frameworks}} \\
DeLLMa & \ding{55} & 24.0 & 39.3 & 0.719 & 0.446 & -- \\
DecisionFlow & \ding{55} & 17.1 & 30.1 & 0.694 & 0.373 & -- \\
Active Pref. Inference & \ding{51} & 9.8 & 18.9 & 0.667 & 0.276 & 9.7 \\
\midrule
\multicolumn{7}{l}{\textsc{Prompting Baselines}} \\
LLM-Direct (B1) & \ding{55} & 62.3 & 83.5 & 0.793 & 0.773 & -- \\
LLM-CoT (B2) & \ding{55} & \cellcolor{lightred}67.0 & \cellcolor{lightred}85.5 & \cellcolor{lightred}0.827 & \cellcolor{lightred}0.803 & -- \\
LLM-Structured Dialogue (B3) & \ding{51} & 13.4 & 26.3 & 0.265 & 0.336 & 4.1 \\
LLM-Free Dialogue (B4) & \ding{51} & 20.2 & 35.2 & 0.364 & 0.410 & 2.3 \\
\midrule
\textbf{\textsc{Decisive} (Ours)} & \ding{51} & \cellcolor{lightgreen}79.0 & \cellcolor{lightgreen}91.6 & \cellcolor{lightgreen}0.887 & \cellcolor{lightgreen}0.893 & 6.1 \\
% \textbf{\textsc{Decisive} (Ours)} & \ding{51} & \cellcolor{lightgreen}73.1 & \cellcolor{lightgreen}85.1 & \cellcolor{lightgreen}0.841 & \cellcolor{lightgreen}0.830 & 4.3 \\
\midrule
\rowcolor{gray!15} \multicolumn{7}{c}{\textbf{Hiring Domain}} \\
\midrule
\multicolumn{7}{l}{\textsc{Decision Frameworks}} \\
DeLLMa & \ding{55} & 9.8 & 20.0 & 0.667 & 0.288 & -- \\
DecisionFlow & \ding{55} & 9.2 & 20.4 & 0.665 & 0.292 & -- \\
Active Pref. Inference & \ding{51} & 32.0 & 56.6 & 0.749 & 0.554 & 7.5 \\
\midrule
\multicolumn{7}{l}{\textsc{Prompting Baselines}} \\
LLM-Direct (B1) & \ding{55} & 87.2 & 97.2 & 0.942 & 0.931 & -- \\
LLM-CoT (B2) & \ding{55} & \cellcolor{lightred}87.8 & \cellcolor{lightgreen}97.8 & \cellcolor{lightred}0.949 & \cellcolor{lightred}0.935 & -- \\
LLM-Structured Dialogue (B3) & \ding{51} & 11.4 & 21.8 & 0.229 & 0.303 & 6.9 \\
LLM-Free Dialogue (B4) & \ding{51} & 31.6 & 40.6 & 0.416 & 0.472 & 2.4 \\
\midrule
\textbf{\textsc{Decisive} (Ours)} & \ding{51} & \cellcolor{lightgreen}90.2 & \cellcolor{lightred}97.6 & \cellcolor{lightgreen}0.958 & \cellcolor{lightgreen}0.947 & 1.9 \\
% \textbf{\textsc{Decisive} (Ours)} & \ding{51} & \cellcolor{lightgreen}89.2 & \cellcolor{lightred}97.4 & \cellcolor{lightgreen}0.953 & \cellcolor{lightgreen}0.942 & 1.53 \\
\bottomrule
\end{tabular}
}
\caption{Performance comparison on Education, Finance, and Hiring domains. We compare \textsc{Decisive} against LLM baselines and existing decision-making frameworks. The ``Dialogue'' column indicates whether the method interactively elicits preferences from the user. \textcolor{lightgreen}{Green} highlights best performance; \textcolor{lightred}{Red} highlights second-best.}
\label{tab:main_results}
\end{table*}

\section{Experimental Setup}
% \AJ{pref vector and user sim add, sampling how many profiles, put sample data in appendix}
% To evaluate our method, we conduct a comprehensive set of experiments comparing it against several strong baselines. 

\subsection{Evaluation Protocol and Metrics}
For each decision scenario, we simulate user interactions by sampling ground-truth preference vectors $\mathbf{w}^*$ from a symmetric Dirichlet distribution. This ensures diverse preference profiles across trials. The ground-truth decision is computed as $o^* = \arg\max_i (\mathbf{S} \mathbf{w}^*)_i$. During interaction, a deterministic user simulator answers tradeoff questions by comparing the corresponding weights: given a question ``$f_a$ vs $f_b$?'', it returns $f_a$ if $w^*_a > w^*_b$, else $f_b$. For our method, we use $P=500$ preference personas sampled from the same Dirichlet prior, which our ablation study (Sec.~\ref{sec:ablation}) shows provides a strong accuracy-efficiency trade-off.

Our evaluation focuses on three aspects: decision accuracy, ranking quality, and interaction efficiency. For decision accuracy, we measure both \textit{Top-1 Accuracy} (percentage of times the top recommendation matches ground truth) and \textit{Top-2 Accuracy}, also known as Hit Rate@2, (ground truth appears in top two). For ranking quality, we use \textit{NDCG@3} and \textit{Mean Reciprocal Rank (MRR)}. For efficiency, we measure \textit{Average Questions} asked before reaching a decision, which serves as a proxy for user effort. 

\subsection{Baselines}
We use two categories of baselines:

\noindent\textbf{Decision Frameworks.} We compare against three recent works: two frameworks for structured decision-making and one for interactive preference elicitation. \textbf{DeLLMa}~\cite{liu2024dellmadecisionmakinguncertainty}: A multi-step reasoning framework that optimizes decisions under uncertainty. We adapt it to our setting by providing the ground-truth preference profile $\mathbf{w}^*$ as part of the context. \textbf{DecisionFlow}~\cite{chen2025decisionflowadvancinglargelanguage}: A system that structures decisions by modeling actions and attributes. Similar to DeLLMa, it is provided with the ground-truth preference profile. \textbf{Active Preference Inference}~\cite{piriyakulkij2024activepreferenceinferenceusing}: An interactive method that enables an LLM to ask questions to infer user preferences. Like our method, it starts without knowledge of $\mathbf{w}^*$ and must elicit it through dialogue.

\noindent\textbf{Prompting Baselines.} These baselines assess the capability of LLMs (GPT-4o) to solve the task directly with varying degrees of assistance. \textbf{LLM-Direct (B1):} A single-shot baseline that receives all inputs (documents, decision question, extracted parameters, and the option-scoring matrix, along with the ground-truth user preference vector $\mathbf{w}^*$). It is prompted to directly output its recommended decision. \textbf{LLM-CoT (B2):} Identical to B1 but utilizes Chain-of-Thought (CoT) prompting to encourage reasoning before the final decision. \textbf{LLM-Structured Dialogue (B3):} An interactive baseline where the LLM does \textit{not} receive $\mathbf{w}^*$. Instead, it must discover user preferences by asking up to 10 yes/no questions before making a final recommendation. This mirrors the setting of our method but relies on the LLM's implicit ability to infer preferences. \textbf{LLM-Free Dialogue (B4):} Similar to B3 but the LLM decides what questions to ask, how many, and when to stop. This tests whether unconstrained dialogue improves preference inference. Exact prompts for all baselines are provided in Appendix~\ref{sec:appendix_prompts}.

% \subsection{Experiments}

% To rigorously assess the effectiveness of Decisive, we designed and executed a series of experiments. The primary goal was to systematically evaluate its performance against both advanced LLM baselines and established academic frameworks, while also validating its alignment with human judgment.

% \subsubsection{Experimental Setup}
% For each of the \(n=150\) decision-making tasks, we quantitatively assessed performance using a comprehensive suite of metrics: \emph{Exact Accuracy} and \emph{Top-2 Soft Accuracy} for output precision; \emph{NDCG (k=3)} and \emph{Mean Reciprocal Rank (MRR)} for ranked-list quality; and the \emph{average number of clarification questions} as a measure of interaction efficiency. 

\section{Results and Discussion}

% \AJ{show ablation of num of profiles, confidence threshold, We should talk about neutral and both also for the paper, right now easy for eval but system can support. Secondly, too many llm calls in baseline, How to come from user pref to pref vector, Our method params not defined I think atleast alpha kitna profile sample kar rahe wo not mentioned, put sample data in appendix}

% Table~\ref{tab:main_results} presents our main results across domains.

% \subsection{Main Results}

\noindent
\textbf{Performance vs. Prompting Methods.}
Given the ground-truth preference vector, LLM baselines (B1, B2) perform reasonably well, achieving $62$-$67\%$ accuracy on Education and Finance, and $\sim$$87\%$ on the Hiring domain where candidate differentiations are more apparent (Table \ref{tab:main_results}). That said, their accuracy is capped by the need to reason over a $10 \times K$ scoring matrix (where $K \approx 11$) and compute weighted preference combinations, which remains challenging due to known LLM limitations in multi-step numerical reasoning. When preferences are unknown (B3, B4), performance collapses across all domains ($9$-$30\%$), confirming that LLMs struggle to implicitly infer and track user preferences through dialogue~\cite{zhao2025llmsrecognizepreferencesevaluating, tsaknakis2025llmsrecognizelatentpreferences}. \textsc{Decisive} addresses this by offloading preference tracking to an explicit Bayesian model, achieving $\sim$$79\%$ on Education and Finance, and $90\%$ on Hiring, \textbf{consistently outperforming} the fully-informed baselines.

\vspace{6pt}
\noindent
\textbf{Performance vs. Decision Frameworks.}
Existing structured frameworks struggle on our benchmark. DeLLMa and DecisionFlow achieve only 9-28\% accuracy despite being provided with the ground-truth preference profile. These frameworks assume a simplified decision structure where a single metric dominates, reducing the decision to a direct optimization problem. Our benchmark, by contrast, requires reasoning over multi-objective trade-offs where users must balance competing factors. 
% This complexity exposes their inability to integrate nuanced preference structures into their decision process.
Active Preference Inference, the only other interactive method, shows highly variable performance (10-32\% across domains), highlighting the limitations of generic entropy-based question selection. Our \textit{decision-aware} strategy, which focuses on factors that differentiate the top options, yields more consistent and substantially higher accuracy.

\subsection{Ablation Studies}
\label{sec:ablation}
\textbf{Effect of Number of Preference Profiles.}
We analyze how the number of sampled preference profiles ($P$) affects accuracy and inference time (Table~\ref{tab:ablation_profiles}). $P$ can be seen as a resolution parameter for our Monte Carlo estimator, where higher values yield a finer-grained approximation of the posterior. Accuracy improves steadily as $P$ increases: from 66.9\% at $P=10$ to 82.7\% at $P=500$. Beyond this point, gains become marginal (83.9\% at $P=2000$) while inference time continues to grow, validating that the EIG estimates at $P=500$ are stable enough to select effective questions. 

\begin{table}[h]
\centering
\scriptsize
\begin{tabular}{c c c c c}
\toprule
\textbf{Profiles ($P$)} & \textbf{Top-1 Acc.} & \textbf{Top-2 Acc.} & \textbf{Time (s)}  &\textbf{Avg. Qs}\\
\midrule
10& 66.9& 84.5& 0.021&2.41\\
100 & 77.5& 89.5&  0.086&3.23\\
500 & 82.7& 93.7&  0.533&4.24\\
1000 & 82.9& 92.6&  0.849&4.01\\
2000 & 83.9& 93.4&  1.289&4.54\\
\bottomrule
\end{tabular}
\caption{Effect of number of preference profiles on accuracy and inference time averaged across all domains.}
\label{tab:ablation_profiles}
\end{table}

% \vspace{-6pt}
We use $P=500$ for our main experiments, which achieves strong accuracy in under 600ms per scenario averaged across all domains.

% \vspace{6pt}
\noindent
\textbf{Robustness to Noisy Responses.}
\label{sec:noise}
To validate robustness to inconsistent user responses, we conduct an ablation using a Bradley-Terry response model where the simulated user occasionally gives inconsistent answers. We compare deterministic responses against noisy responses (temperature = 0.05) across all three domains (Table~\ref{tab:noise_ablation}). 

\begin{table}[h]
\centering
\tiny
\begin{tabular}{l l c c c c c}
\toprule
\textbf{Temp} & \textbf{Domain} & \textbf{Top-1} & \textbf{Top-2} & \textbf{NDCG@3} & \textbf{MRR} & \textbf{Avg Qs} \\
\midrule
None & Education & 78.8 & 91.6 & 0.893 & 0.879 & 5.1 \\
None & Finance & 79.0 & 91.6 & 0.887 & 0.861 & 6.1 \\
None & Hiring & 90.2 & 97.8 & 0.958 & 0.947 & 1.9 \\
\midrule
0.05 & Education & 76.6 & 91.0 & 0.881 & 0.866 & 5.5 \\
0.05 & Finance & 76.8 & 90.4 & 0.880 & 0.861 & 6.2 \\
0.05 & Hiring & 89.6 & 97.4 & 0.955 & 0.943 & 1.8 \\
\bottomrule
\end{tabular}
\caption{Noise robustness ablation. Performance under deterministic (Temp=None) vs.\ noisy (Temp=0.05) Bradley-Terry user responses.}
\label{tab:noise_ablation}
\end{table}

% \vspace{-4pt}
Performance remains largely stable under noise: average Top-1 drops only $\sim$2-2.5\% and Top-2 drops $\sim$1\%, with Hiring remaining particularly robust. These results confirm that even when user responses deviate from the model's assumptions, the soft reweighting mechanism prevents catastrophic information loss.

% \vspace{6pt}
\noindent
\textbf{Extensibility to Nuanced Responses.}
A key advantage of our probabilistic formulation is its extensibility to nuanced user feedback. Because we model preferences as distributions over weight vectors, the framework can naturally accommodate responses beyond simple binary choices. For instance, when a user indicates two factors are \textit{equally important} or they don't have any strong preference for either (neutral), the update rule can preserve the prior over those factors. When \textit{both} are highly valued, both can be upweighted proportionally. This flexibility is not possible in methods like Active Preference Inference, which rely on hard binary elimination. 
% and cannot accommodate such responses without ad-hoc modifications.

\subsection{Computational Efficiency}
\textbf{Question Efficiency.}
\textsc{Decisive} requires only 5-6 questions on average to reach a confident decision in Education and Finance, and under 2 questions in Hiring where candidates are more clearly differentiated. This is fewer than Active Preference Inference ($\sim$7.5-9.7 questions) and comparable to B3 ($\sim$4-7 questions), but with substantially higher accuracy. Our dynamic stopping criterion terminates the dialogue only when the \textit{decision distribution} is confident, avoiding wasted effort on preferences that do not impact the final choice.\\
\noindent
\textbf{LLM Call Efficiency.}
% An important practical consideration is the number of LLM API calls required per decision. 
DecisionFlow's multi-step pipeline (extraction, attribute mapping, weight computation per option-attribute pair) requires dozens of LLM calls per scenario. DeLLMa and the dialogue baselines (B3, B4, Active Preference) each require 7-10 calls. In contrast, \textsc{Decisive} requires LLM calls only for initial parameter extraction and Option-Scoring Matrix construction. The subsequent Bayesian inference and question selection phases are purely computational, requiring \textbf{zero LLM calls} during user interaction. This design makes our approach significantly more efficient and cost-effective at inference time.

\section{Conclusion}
We present \textsc{Decisive}, a framework that transforms the passive retrieval of unstructured information into an active, personalized decision-making process. By explicitly decoupling factual evidence extraction from subjective preference modeling, our approach addresses the twin challenges of information overload and preference ambiguity. Our results reveal three key findings: (1) Structured preference integration significantly improves accuracy over text-based reasoning. (2) Dialogue alone is insufficient for preference discovery without explicit state tracking. (3) Decision-aware elicitation enables efficient convergence with fewer questions than entropy-based methods. We believe \textsc{Decisive} represents a step toward next-generation decision engines that do not merely present information, but actively collaborate with users to assist them in complex, real-world decision-making scenarios. 
% (1) \textit{Structured preference integration matters}: B1/B2 with the option-scoring matrix outperform other baselines by 2--3$\times$, even though all receive $w^*$ directly. 
% (2) \textit{Dialogue alone is insufficient}: B3 and Active Preference achieve only 9--20\% accuracy despite multi-turn interaction, confirming that LLMs cannot implicitly perform Bayesian preference inference. 
% (3) \textit{Decision-aware elicitation is critical}: Our information-gain-driven question selection enables efficient preference discovery with fewer questions than Active Preference while achieving substantially higher accuracy.

% Our results demonstrate that this structured, interactive approach significantly outperforms both general-purpose LLMs and existing decision frameworks, achieving high decision accuracy with minimal user effort. We believe \textsc{Decisive} represents a step toward next-generation decision engines that do not merely present information, but actively collaborate with users to assist them in complex, real-world decision-making scenarios. 

\section*{Limitations}
Our framework relies on the quality of the Option-Scoring Matrix, which depends on the extraction capabilities of the underlying LLM. Errors or hallucinations in scoring can propagate to the final decision; we mitigate this by aggregating scores from multiple models, but residual noise may still persist. Additionally, our current utility model assumes that decision factors are independent. Real-world preferences can exhibit interdependencies (e.g., a user might prioritize quality only if price falls below a certain threshold), which a linear model may not fully capture. The framework also extracts a fixed set of $K$ factors per scenario; when users have goals that fall outside the extracted factors (e.g., personal connections or emotional considerations not reflected in documents), the system's recommendations may not fully capture their preferences. Finally, our evaluation uses simulated users with known ground-truth preferences. While we validate robustness under noisy responses via a Bradley-Terry ablation (Section~\ref{sec:noise}),aspects such as usability and user satisfaction remain to be validated through human studies.

\section*{Ethics Statement}
We recognize that automated decision support, particularly in high-stakes domains like hiring, finance, and education, carries significant ethical responsibilities. \textsc{Decisive} is designed as an assistive tool to augment, not replace, human agency. By explicitly separating objective evidence extraction from subjective preference modeling, our framework aims to increase transparency and allow users to understand the basis of recommendations. However, we acknowledge that the underlying LLMs used for option scoring may exhibit latent biases, which could propagate to the decision process. While we mitigate this by aggregating scores across multiple models, users should interpret recommendations as data-driven suggestions rather than definitive judgments. Furthermore, all data used in our experiments is either synthetically generated or sourced from public, anonymized repositories (e.g., Kaggle resumes), ensuring that no private individual data was compromised.

\bibliography{anthology,custom}

@misc{liu2024dellmadecisionmakinguncertainty,
      title={DeLLMa: Decision Making Under Uncertainty with Large Language Models}, 
      author={Ollie Liu and Deqing Fu and Dani Yogatama and Willie Neiswanger},
      year={2024},
      eprint={2402.02392},
      archivePrefix={arXiv},
      primaryClass={cs.AI},
      url={https://arxiv.org/abs/2402.02392}, 
}

@misc{chen2025decisionflowadvancinglargelanguage,
      title={DecisionFlow: Advancing Large Language Model as Principled Decision Maker}, 
      author={Xiusi Chen and Shanyong Wang and Cheng Qian and Hongru Wang and Peixuan Han and Heng Ji},
      year={2025},
      eprint={2505.21397},
      archivePrefix={arXiv},
      primaryClass={cs.CL},
      url={https://arxiv.org/abs/2505.21397}, 
}

@misc{hu2024languagemodelsalignabledecisionmakers,
      title={Language Models are Alignable Decision-Makers: Dataset and Application to the Medical Triage Domain}, 
      author={Brian Hu and Bill Ray and Alice Leung and Amy Summerville and David Joy and Christopher Funk and Arslan Basharat},
      year={2024},
      eprint={2406.06435},
      archivePrefix={arXiv},
      primaryClass={cs.CL},
      url={https://arxiv.org/abs/2406.06435}, 
}

@misc{nazi2024largelanguagemodelshealthcare,
      title={Large language models in healthcare and medical domain: A review}, 
      author={Zabir Al Nazi and Wei Peng},
      year={2024},
      eprint={2401.06775},
      archivePrefix={arXiv},
      primaryClass={cs.CL},
      url={https://arxiv.org/abs/2401.06775}, 
}

@article{LI20251,
title = {Large language models-powered clinical decision support: enhancing or replacing human expertise?},
journal = {Intelligent Medicine},
volume = {5},
number = {1},
pages = {1-4},
year = {2025},
issn = {2667-1026},
doi = {https://doi.org/10.1016/j.imed.2025.01.001},
url = {https://www.sciencedirect.com/science/article/pii/S2667102625000014},
author = {Jia Li and Zichun Zhou and Han Lyu and Zhenchang Wang}
}

@article{maity2025llmhealthcare,
  author    = {Maity, S. and Saikia, M. J.},
  title     = {Large Language Models in Healthcare and Medical Applications: A Review},
  journal   = {Bioengineering (Basel)},
  year      = {2025},
  volume    = {12},
  number    = {6},
  pages     = {631},
  doi       = {10.3390/bioengineering12060631},
  pmid      = {40564447},
  pmcid     = {PMC12189880},
  publisher = {MDPI},
  month     = jun,
  day       = {10}
}

@misc{kim2024mdagentsadaptivecollaborationllms,
      title={MDAgents: An Adaptive Collaboration of LLMs for Medical Decision-Making}, 
      author={Yubin Kim and Chanwoo Park and Hyewon Jeong and Yik Siu Chan and Xuhai Xu and Daniel McDuff and Hyeonhoon Lee and Marzyeh Ghassemi and Cynthia Breazeal and Hae Won Park},
      year={2024},
      eprint={2404.15155},
      archivePrefix={arXiv},
      primaryClass={cs.CL},
      url={https://arxiv.org/abs/2404.15155}, 
}

@article{GUMILAR20244019,
title = {Assessment of Large Language Models (LLMs) in decision-making support for gynecologic oncology},
journal = {Computational and Structural Biotechnology Journal},
volume = {23},
pages = {4019-4026},
year = {2024},
issn = {2001-0370},
doi = {https://doi.org/10.1016/j.csbj.2024.10.050},
url = {https://www.sciencedirect.com/science/article/pii/S2001037024003702},
author = {Khanisyah Erza Gumilar and Birama R. Indraprasta and Ach Salman Faridzi and Bagus M. Wibowo and Aditya Herlambang and Eccita Rahestyningtyas and Budi Irawan and Zulkarnain Tambunan and Ahmad Fadhli Bustomi and Bagus Ngurah Brahmantara and Zih-Ying Yu and Yu-Cheng Hsu and Herlangga Pramuditya and Very Great E. Putra and Hari Nugroho and Pungky Mulawardhana and Brahmana A. Tjokroprawiro and Tri Hedianto and Ibrahim H. Ibrahim and Jingshan Huang and Dongqi Li and Chien-Hsing Lu and Jer-Yen Yang and Li-Na Liao and Ming Tan}
}

@inproceedings{10.1145/3715336.3735779,
author = {Foo, Charisse and Foong, Pin Sym and Nadal, Camille and Ureyang, Natasha and Naylin, Thant and Koh, Gerald Choon Huat},
title = {The Benefits and Risks of LLMs for Facilitating Medical Decision-Making Among Laypersons},
year = {2025},
isbn = {9798400714856},
publisher = {Association for Computing Machinery},
address = {New York, NY, USA},
url = {https://doi.org/10.1145/3715336.3735779},
doi = {10.1145/3715336.3735779},
booktitle = {Proceedings of the 2025 ACM Designing Interactive Systems Conference},
pages = {3173–3191},
numpages = {19},
location = {
},
series = {DIS '25}
}

@misc{eigner2024determinantsllmassisteddecisionmaking,
      title={Determinants of LLM-assisted Decision-Making}, 
      author={Eva Eigner and Thorsten Händler},
      year={2024},
      eprint={2402.17385},
      archivePrefix={arXiv},
      primaryClass={cs.AI},
      url={https://arxiv.org/abs/2402.17385}, 
}

@article{changeux2024strategic,
  author    = {Alexander Changeux and Stephen Montagnier},
  title     = {Strategic Decision-Making Support Using Large Language Models (LLMs)},
  journal   = {Management Journal for Advanced Research},
  year      = {2024},
  volume    = {4},
  number    = {4},
  pages     = {102--108},
  month     = aug,
  doi       = {10.5281/zenodo.13444483},
  url       = {https://mjar.singhpublication.com/index.php/ojs/article/view/160}
}

@misc{simchilevi2025largelanguagemodelssupply,
      title={Large Language Models for Supply Chain Decisions}, 
      author={David Simchi-Levi and Konstantina Mellou and Ishai Menache and Jeevan Pathuri},
      year={2025},
      eprint={2507.21502},
      archivePrefix={arXiv},
      primaryClass={cs.AI},
      url={https://arxiv.org/abs/2507.21502}, 
}

@misc{li2024investorbenchbenchmarkfinancialdecisionmaking,
      title={INVESTORBENCH: A Benchmark for Financial Decision-Making Tasks with LLM-based Agent}, 
      author={Haohang Li and Yupeng Cao and Yangyang Yu and Shashidhar Reddy Javaji and Zhiyang Deng and Yueru He and Yuechen Jiang and Zining Zhu and Koduvayur Subbalakshmi and Guojun Xiong and Jimin Huang and Lingfei Qian and Xueqing Peng and Qianqian Xie and Jordan W. Suchow},
      year={2024},
      eprint={2412.18174},
      archivePrefix={arXiv},
      primaryClass={cs.CE},
      url={https://arxiv.org/abs/2412.18174}, 
}

@misc{yu2024finconsynthesizedllmmultiagent,
      title={FinCon: A Synthesized LLM Multi-Agent System with Conceptual Verbal Reinforcement for Enhanced Financial Decision Making}, 
      author={Yangyang Yu and Zhiyuan Yao and Haohang Li and Zhiyang Deng and Yupeng Cao and Zhi Chen and Jordan W. Suchow and Rong Liu and Zhenyu Cui and Zhaozhuo Xu and Denghui Zhang and Koduvayur Subbalakshmi and Guojun Xiong and Yueru He and Jimin Huang and Dong Li and Qianqian Xie},
      year={2024},
      eprint={2407.06567},
      archivePrefix={arXiv},
      primaryClass={cs.CL},
      url={https://arxiv.org/abs/2407.06567}, 
}

@misc{andukuri2024stargateteachinglanguagemodels,
      title={STaR-GATE: Teaching Language Models to Ask Clarifying Questions}, 
      author={Chinmaya Andukuri and Jan-Philipp Fränken and Tobias Gerstenberg and Noah D. Goodman},
      year={2024},
      eprint={2403.19154},
      archivePrefix={arXiv},
      primaryClass={cs.CL},
      url={https://arxiv.org/abs/2403.19154}, 
}

@misc{dou2025togateclarifyingquestionssummarizing,
      title={TO-GATE: Clarifying Questions and Summarizing Responses with Trajectory Optimization for Eliciting Human Preference}, 
      author={Yulin Dou and Jiangming Liu},
      year={2025},
      eprint={2506.02827},
      archivePrefix={arXiv},
      primaryClass={cs.CL},
      url={https://arxiv.org/abs/2506.02827}, 
}

@misc{piriyakulkij2024activepreferenceinferenceusing,
      title={Active Preference Inference using Language Models and Probabilistic Reasoning}, 
      author={Wasu Top Piriyakulkij and Volodymyr Kuleshov and Kevin Ellis},
      year={2024},
      eprint={2312.12009},
      archivePrefix={arXiv},
      primaryClass={cs.CL},
      url={https://arxiv.org/abs/2312.12009}, 
}

@misc{park2025choicematessupportingunfamiliaronline,
      title={ChoiceMates: Supporting Unfamiliar Online Decision-Making with Multi-Agent Conversational Interactions}, 
      author={Jeongeon Park and Bryan Min and Kihoon Son and Jean Y. Song and Xiaojuan Ma and Juho Kim},
      year={2025},
      eprint={2310.01331},
      archivePrefix={arXiv},
      primaryClass={cs.HC},
      url={https://arxiv.org/abs/2310.01331}, 
}

@misc{jia2024decisionmakingbehaviorevaluationframework,
      title={Decision-Making Behavior Evaluation Framework for LLMs under Uncertain Context}, 
      author={Jingru Jia and Zehua Yuan and Junhao Pan and Paul E. McNamara and Deming Chen},
      year={2024},
      eprint={2406.05972},
      archivePrefix={arXiv},
      primaryClass={cs.AI},
      url={https://arxiv.org/abs/2406.05972}, 
}

@inproceedings{zhang-choi-2025-clarify,
    title = "Clarify When Necessary: Resolving Ambiguity Through Interaction with {LM}s",
    author = "Zhang, Michael JQ  and
      Choi, Eunsol",
    editor = "Chiruzzo, Luis  and
      Ritter, Alan  and
      Wang, Lu",
    booktitle = "Findings of the Association for Computational Linguistics: NAACL 2025",
    month = apr,
    year = "2025",
    address = "Albuquerque, New Mexico",
    publisher = "Association for Computational Linguistics",
    url = "https://aclanthology.org/2025.findings-naacl.306/",
    doi = "10.18653/v1/2025.findings-naacl.306",
    pages = "5526--5543",
    ISBN = "979-8-89176-195-7"
}

@misc{chang_teaching_2025,
	title = {Teaching {AI} to {Clarify}: {Handling} {Assumptions} and {Ambiguity} in {Language} {Models}},
	url = {https://www.shanechang.com/p/training-llms-smarter-clarifying-ambiguity-assumptions/},
	author = {Chang, Shane},
	month = may,
	year = {2025},
}

@misc{deoliveira2025infoquestevaluatingmultiturndialogue,
      title={InfoQuest: Evaluating Multi-Turn Dialogue Agents for Open-Ended Conversations with Hidden Context}, 
      author={Bryan L. M. de Oliveira and Luana G. B. Martins and Bruno Brandão and Luckeciano C. Melo},
      year={2025},
      eprint={2502.12257},
      archivePrefix={arXiv},
      primaryClass={cs.CL},
      url={https://arxiv.org/abs/2502.12257}, 
}

@misc{tsaknakis2025llmsrecognizelatentpreferences,
      title={Do LLMs Recognize Your Latent Preferences? A Benchmark for Latent Information Discovery in Personalized Interaction}, 
      author={Ioannis Tsaknakis and Bingqing Song and Shuyu Gan and Dongyeop Kang and Alfredo Garcia and Gaowen Liu and Charles Fleming and Mingyi Hong},
      year={2025},
      eprint={2510.17132},
      archivePrefix={arXiv},
      primaryClass={cs.LG},
      url={https://arxiv.org/abs/2510.17132}, 
}

@misc{montazeralghaem2025askingclarifyingquestionspreference,
      title={Asking Clarifying Questions for Preference Elicitation With Large Language Models}, 
      author={Ali Montazeralghaem and Guy Tennenholtz and Craig Boutilier and Ofer Meshi},
      year={2025},
      eprint={2510.12015},
      archivePrefix={arXiv},
      primaryClass={cs.AI},
      url={https://arxiv.org/abs/2510.12015}, 
}

@misc{handa2024bayesianpreferenceelicitationlanguage,
      title={Bayesian Preference Elicitation with Language Models}, 
      author={Kunal Handa and Yarin Gal and Ellie Pavlick and Noah Goodman and Jacob Andreas and Alex Tamkin and Belinda Z. Li},
      year={2024},
      eprint={2403.05534},
      archivePrefix={arXiv},
      primaryClass={cs.CL},
      url={https://arxiv.org/abs/2403.05534}, 
}

@inproceedings{10.1145/3706598.3713423,
author = {Ma, Shuai and Chen, Qiaoyi and Wang, Xinru and Zheng, Chengbo and Peng, Zhenhui and Yin, Ming and Ma, Xiaojuan},
title = {Towards Human-AI Deliberation: Design and Evaluation of LLM-Empowered Deliberative AI for AI-Assisted Decision-Making},
year = {2025},
isbn = {9798400713941},
publisher = {Association for Computing Machinery},
address = {New York, NY, USA},
url = {https://doi.org/10.1145/3706598.3713423},
doi = {10.1145/3706598.3713423},
booktitle = {Proceedings of the 2025 CHI Conference on Human Factors in Computing Systems},
articleno = {261},
numpages = {23},
location = {
},
series = {CHI '25}
}

@article{ALBASHRAWI2025100751,
title = {Generative AI for decision-making: A multidisciplinary perspective},
journal = {Journal of Innovation \& Knowledge},
volume = {10},
number = {4},
pages = {100751},
year = {2025},
issn = {2444-569X},
doi = {https://doi.org/10.1016/j.jik.2025.100751},
url = {https://www.sciencedirect.com/science/article/pii/S2444569X25000964},
author = {Mousa Albashrawi}
}

@article{yin_responsible_2025,
	title = {Responsible {AI} in student management: preventing misdecision in career choice of university students under inaccurate guidance},
	volume = {15},
	issn = {2045-2322},
	shorttitle = {Responsible {AI} in student management},
	url = {https://www.nature.com/articles/s41598-025-22127-7},
	doi = {10.1038/s41598-025-22127-7},
	language = {en},
	number = {1},
	urldate = {2026-01-03},
	journal = {Scientific Reports},
	author = {Yin, Shi and Imran, Raiha and Ullah, Kifayat and Ali, Zeeshan and Haleemzai, Izatmand},
	month = oct,
	year = {2025},
	pages = {38177},
}

@inproceedings{wang-etal-2024-jobfair,
    title = "{J}ob{F}air: A Framework for Benchmarking Gender Hiring Bias in Large Language Models",
    author = "Wang, Ze  and
      Wu, Zekun  and
      Guan, Xin  and
      Thaler, Michael  and
      Koshiyama, Adriano  and
      Lu, Skylar  and
      Beepath, Sachin  and
      Ertekin, Ediz  and
      Perez-Ortiz, Maria",
    editor = "Al-Onaizan, Yaser  and
      Bansal, Mohit  and
      Chen, Yun-Nung",
    booktitle = "Findings of the Association for Computational Linguistics: EMNLP 2024",
    month = nov,
    year = "2024",
    address = "Miami, Florida, USA",
    publisher = "Association for Computational Linguistics",
    url = "https://aclanthology.org/2024.findings-emnlp.184/",
    doi = "10.18653/v1/2024.findings-emnlp.184",
    pages = "3227--3246",
}

@misc{veldanda2023emilygregemployablelakisha,
      title={Are Emily and Greg Still More Employable than Lakisha and Jamal? Investigating Algorithmic Hiring Bias in the Era of ChatGPT}, 
      author={Akshaj Kumar Veldanda and Fabian Grob and Shailja Thakur and Hammond Pearce and Benjamin Tan and Ramesh Karri and Siddharth Garg},
      year={2023},
      eprint={2310.05135},
      archivePrefix={arXiv},
      primaryClass={cs.CL},
      url={https://arxiv.org/abs/2310.05135}, 
}

@misc{bhawal2021resume,
  author       = {Snehaan Bhawal},
  title        = {Resume Dataset},
  year         = {2021},
  url          = {https://www.kaggle.com/datasets/snehaanbhawal/resume-dataset}
}

@misc{zhao2025llmsrecognizepreferencesevaluating,
      title={Do LLMs Recognize Your Preferences? Evaluating Personalized Preference Following in LLMs}, 
      author={Siyan Zhao and Mingyi Hong and Yang Liu and Devamanyu Hazarika and Kaixiang Lin},
      year={2025},
      eprint={2502.09597},
      archivePrefix={arXiv},
      primaryClass={cs.LG},
      url={https://arxiv.org/abs/2502.09597}, 
}

@misc{upadhyay2024umbrelaumbrelaopensourcereproduction,
      title={UMBRELA: UMbrela is the (Open-Source Reproduction of the) Bing RELevance Assessor}, 
      author={Shivani Upadhyay and Ronak Pradeep and Nandan Thakur and Nick Craswell and Jimmy Lin},
      year={2024},
      eprint={2406.06519},
      archivePrefix={arXiv},
      primaryClass={cs.IR},
      url={https://arxiv.org/abs/2406.06519}, 
}

@incollection{warwick37961,
            year = {2011},
          series = {Oxford handbooks in mathematics},
           pages = {656--705},
       booktitle = {The Oxford handbook of nonlinear filtering },
           title = {A tutorial on particle filtering and smoothing : fiteen years later
},
          editor = {Dan Crisan and Boris  Rozovskii},
       publisher = {Oxford University Press},
         address = {Oxford ; N.Y. },
          author = {Doucet, Arnaud and Johansen, Adam M.},
             url = {http://webcat.warwick.ac.uk/record=b2490036~S1},
            isbn = {9780199532902}
}

@article{Arulampalam2002ATO,
  title={A tutorial on particle filters for online nonlinear/non-Gaussian Bayesian tracking},
  author={M. Sanjeev Arulampalam and Simon Maskell and Neil J. Gordon and Tim Clapp},
  journal={IEEE Trans. Signal Process.},
  year={2002},
  volume={50},
  pages={174-188},
  url={https://api.semanticscholar.org/CorpusID:55577025}
}

@article{10.1609/aimag.v29i4.2200,
author = {Pu, Pearl and Chen, Li},
title = {User‐Involved Preference Elicitation for Product Search and Recommender Systems},
year = {2008},
issue_date = {Winter 2008},
publisher = {John Wiley \& Sons, Inc.},
address = {USA},
volume = {29},
number = {4},
issn = {0738-4602},
url = {https://doi.org/10.1609/aimag.v29i4.2200},
doi = {10.1609/aimag.v29i4.2200},
journal = {AI Mag.},
month = dec,
pages = {93–103},
numpages = {11}
}

@inproceedings{10.1145/1064009.1064038,
author = {Pu, Pearl and Chen, Li},
title = {Integrating tradeoff support in product search tools for e-commerce sites},
year = {2005},
isbn = {1595930493},
publisher = {Association for Computing Machinery},
address = {New York, NY, USA},
url = {https://doi.org/10.1145/1064009.1064038},
doi = {10.1145/1064009.1064038},
booktitle = {Proceedings of the 6th ACM Conference on Electronic Commerce},
pages = {269–278},
numpages = {10},
location = {Vancouver, BC, Canada},
series = {EC '05}
}

@InProceedings{pmlr-v9-guo10b,
  title = 	 {Real-time Multiattribute Bayesian Preference Elicitation with Pairwise Comparison Queries},
  author = 	 {Guo, Shengbo and Sanner, Scott},
  booktitle = 	 {Proceedings of the Thirteenth International Conference on Artificial Intelligence and Statistics},
  pages = 	 {289--296},
  year = 	 {2010},
  editor = 	 {Teh, Yee Whye and Titterington, Mike},
  volume = 	 {9},
  series = 	 {Proceedings of Machine Learning Research},
  address = 	 {Chia Laguna Resort, Sardinia, Italy},
  month = 	 {13--15 May},
  publisher =    {PMLR},
  url = 	 {https://proceedings.mlr.press/v9/guo10b.html}
}

@inproceedings{10.1145/3640457.3688142,
author = {Austin, David and Korikov, Anton and Toroghi, Armin and Sanner, Scott},
title = {Bayesian Optimization with LLM-Based Acquisition Functions for Natural Language Preference Elicitation},
year = {2024},
isbn = {9798400705052},
publisher = {Association for Computing Machinery},
address = {New York, NY, USA},
url = {https://doi.org/10.1145/3640457.3688142},
doi = {10.1145/3640457.3688142},
booktitle = {Proceedings of the 18th ACM Conference on Recommender Systems},
pages = {74–83},
numpages = {10},
location = {Bari, Italy},
series = {RecSys '24}
}
\bibliographystyle{acl_natbib}

\appendix

\section{Appendix}
\label{sec:appendix}
\subsection{Document Generation Pipeline Details}
\label{sec:appendix_generation}
Our context-aware document synthesis (Stage 2) proceeds in three steps for each document:

\begin{enumerate}[noitemsep,topsep=0pt,leftmargin=*]
\item \textbf{Blueprint Generation:} The system generates a hierarchical JSON blueprint outlining the document's structure, including its sections and subsections (e.g., Program Overview, Fee Structure, Career Prospects for a university prospectus).

\item \textbf{Sequential Content Generation:} Content is generated section by section, where each new section is explicitly conditioned on all previously generated sections. This sequential conditioning maintains factual consistency (e.g., ensuring fee amounts mentioned in one section match those in another) and stylistic coherence across the document.

\item \textbf{Compilation and Refinement:} All generated sections are compiled, and a concluding LLM call refines the complete document to ensure stylistic uniformity, remove generation artifacts, and verify internal consistency.
\end{enumerate}

\noindent This hierarchical approach ensures that generated documents exhibit the coherence characteristics of real-world documents, avoiding the inconsistencies that often arise from single-shot generation.

\subsection{Baseline Prompts}
\label{sec:appendix_prompts}

Below we provide the exact prompts used for each prompting baseline. In all prompts, \texttt{<QUESTION>}, \texttt{<OPTIONS>}, \texttt{<PARAMS>}, \texttt{<G\_MATRIX>}, \texttt{<DOCUMENTS>}, and \texttt{<USER\_PREFS>} are placeholders populated from the scenario data.

\subsubsection{B1: LLM-Direct}
\begin{framed}\scriptsize
\begin{verbatim}
You are a decision-making assistant.

=== DECISION QUESTION ===
<QUESTION>

=== OPTIONS ===
<OPTIONS>

=== USER PREFERENCES ===
<USER_PREFS>

=== PARAMETERS ===
<PARAMS>

=== G-MATRIX (How each option scores on each
    parameter, 0-1 scale) ===
<G_MATRIX>

=== DOCUMENTS ===
<DOCUMENTS>

=== TASK ===
Based on the user's preferences and the
information above, select the best option
for this user.

OUTPUT FORMAT (JSON only):
{
  "ranking": ["best_option", "second_best", ...],
  "decision": "best_option"
}
\end{verbatim}
\end{framed}

\subsubsection{B2: LLM-CoT}
Identical to B1, with the following task section replaced:
\begin{framed}\scriptsize
\begin{verbatim}
=== TASK ===
Based on the user's preferences and the
information provided, determine which option
is best for this user.

Let's think step by step:
- First, identify which parameters matter most
  to this user based on their preference weights.
- Then, analyze how each option scores on those
  key parameters.
- Finally, weigh the trade-offs and make your
  decision.

Respond with JSON in this format:
{
  "reasoning": "Your step-by-step analysis...",
  "ranking": ["best_option", "second_best", ...],
  "decision": "best_option"
}
\end{verbatim}
\end{framed}

\subsubsection{B3: LLM-Structured Dialogue}
The LLM receives the decision context (question, options, parameters, G-matrix, abbreviated documents) but \textit{not} the user's preference vector. It is instructed to ask up to 10 yes/no questions:
\begin{framed}\scriptsize
\begin{verbatim}
You are a helpful decision assistant having a
conversation with a user.

=== CONTEXT ===
Decision Question: <QUESTION>
Options: <OPTIONS>
Parameters that matter: <PARAMS>
G-Matrix: <G_MATRIX>
Documents (abbreviated): <DOCUMENTS>

=== YOUR TASK ===
You need to help the user choose the best option.
To do this:
1. Ask the user YES/NO questions to understand
   their preferences
2. After gathering enough information (3-5
   questions), make a recommendation

=== RULES ===
- Ask ONE question at a time
- Questions must be answerable with YES or NO
- Focus on understanding which parameters matter
  most to the user
- When ready to recommend, provide your decision

=== OUTPUT FORMAT ===
For each turn, output JSON:
{
  "type": "question" or "decision",
  "content": "Your question" (if asking),
  "ranking": ["best", "2nd best", ...] (if
    deciding),
  "reasoning": "Brief reasoning"
}
\end{verbatim}
\end{framed}

\subsubsection{B4: LLM-Free Dialogue}
Similar to B3, but the LLM has full autonomy over question format, count, and stopping:
\begin{framed}\scriptsize
\begin{verbatim}
You are a decision assistant helping a user
make a choice.

=== DECISION CONTEXT ===
Question: <QUESTION>
Options: <OPTIONS>
Relevant Factors: <PARAMS>
How options score on these factors: <G_MATRIX>
Documents (abbreviated): <DOCUMENTS>

=== YOUR TASK ===
Help the user make the best decision for THEIR
specific needs.

You have FULL AUTONOMY over your strategy:
- You may ask the user questions to understand
  their preferences
- You may ask ANY type of question (open-ended,
  comparison, yes/no, etc.)
- You decide how many questions to ask (could
  be 0 if you think you have enough info)
- You decide when you're ready to make a
  recommendation

The user knows their own preferences but hasn't
studied the options in detail. Your job is to
match their preferences to the best option.

=== OUTPUT FORMAT ===
Respond with JSON:
{
  "action": "ask" or "decide",
  "content": "Your question" (if asking),
  "ranking": ["best", "2nd best", ...] (if
    deciding),
  "reasoning": "Why you're asking / why this
    recommendation"
}
\end{verbatim}
\end{framed}

\subsection{Candidate Set Size Ablation}
\label{sec:appendix_m5}

We ablate the effect of candidate set size by reducing $M$ from 10 to 5, comparing \textsc{Decisive} against the two strongest baselines (B1 and B2) on a representative subset (Table~\ref{tab:m5_ablation}).
\begin{table}[h]
\centering
\tiny
\begin{tabular}{l l c c c c c}
\toprule
 \textbf{Domain} & \textbf{Method} & \textbf{Top-1} & \textbf{Top-2} & \textbf{NDCG@3} & \textbf{MRR} & \textbf{Avg Qs} \\
\midrule
 Education & \textsc{Decisive} & 81.0 & 93.8 & 0.930 & 0.881 & 3.7 \\
 Education & B1 & 75.0 & 91.0 & 0.908 & 0.856 & -- \\
 Education & B2 & 78.2 & 89.0 & 0.909 & 0.881 & -- \\
\midrule
 Finance & \textsc{Decisive} & 79.0 & 95.0 & 0.922 & 0.878 & 3.4 \\
 Finance & B1 & 76.0 & 91.0 & 0.911 & 0.872 & -- \\
 Finance & B2 & 77.4 & 92.3 & 0.896 & 0.873 & -- \\
\midrule
 Hiring & \textsc{Decisive} & 93.4 & 98.0 & 0.974 & 0.964 & 1.2 \\
 Hiring & B1 & 91.0 & 97.4 & 0.967 & 0.955 & -- \\
 Hiring & B2 & 92.0 & 97.6 & 0.970 & 0.958 & -- \\
\bottomrule
\end{tabular}
\caption{Effect of candidate set size ($M$=5). \textsc{Decisive} outperforms both fully-informed baselines across all domains, with fewer interaction questions required compared to $M$=10.}
\label{tab:m5_ablation}
\end{table}

 \textsc{Decisive} outperforms both baselines even at $M$=5 while requiring fewer questions. This is expected because our preference elicitation operates over decision factors rather than individual options, allowing the framework to scale naturally with $M$. In contrast, prompting-based baselines must fit the entire option set and associated documents into the context window, which becomes increasingly limiting as $M$ grows. Notably, the gains over prompting baselines are more pronounced at larger $M$ (Table~\ref{tab:main_results}), where simple prompting-based approaches struggle to jointly reason over all candidates.

\end{document}